\documentclass{article}
\usepackage[accepted]{inn2020}

\usepackage{microtype}
\usepackage{graphicx}
\usepackage{booktabs}

\usepackage{times}

\usepackage{hyperref}
\usepackage{url}
\usepackage{booktabs}

\usepackage[english]{babel}

\usepackage[utf8x]{inputenc}
\usepackage[T1]{fontenc}
\usepackage[noend]{algpseudocode}
\usepackage{fancyhdr,amssymb}
\usepackage[ruled,vlined]{algorithm2e}
\usepackage{array}
\usepackage{stmaryrd}
\usepackage{amsfonts}

\usepackage{amsfonts,amsmath,amssymb,amsthm,epsfig,psfrag}
\usepackage[T1]{fontenc}
\usepackage{babel}

\usepackage{amsmath}
\usepackage{mathtools}
\usepackage{dsfont}
\usepackage{systeme}
\usepackage{graphicx}
\usepackage[colorinlistoftodos]{todonotes}
\usepackage{blindtext}
\usepackage{enumitem}
\usepackage{xcolor}
\usepackage{lipsum}
\usepackage[export]{adjustbox}

\usepackage{array}
\newcommand{\PreserveBackslash}[1]{\let\temp=\\#1\let\\=\temp}
\newcolumntype{C}[1]{>{\PreserveBackslash\centering}p{#1}}
\newcolumntype{R}[1]{>{\PreserveBackslash\raggedleft}p{#1}}
\newcolumntype{L}[1]{>{\PreserveBackslash\raggedright}p{#1}}

\newcommand{\cmmnt}[1]{}
\usepackage{tikz}

\newcolumntype{M}[1]{>{\centering\arraybackslash}m{#1}}
\newcolumntype{N}{@{}m{0pt}@{}}

\setlist[description]{leftmargin=\parindent,labelindent=\parindent}

\icmltitlerunning{Time series source separation with slow flows}

\begin{document}

\twocolumn[
\icmltitle{Time Series Source Separation with Slow Flows}

\begin{icmlauthorlist}
\icmlauthor{Edouard Pineau}{to,goo}
\icmlauthor{Sébastien Razakarivony}{to}
\icmlauthor{Thomas Bonald}{goo}
\end{icmlauthorlist}

\icmlaffiliation{to}{Safran Tech, Signal and Information Technologies}
\icmlaffiliation{goo}{Telecom Paris, Institut Polytechnique de Paris}

\icmlcorrespondingauthor{Edouard Pineau}{pineau.edouard@gmail.com}

\icmlkeywords{Time series, Normalizing flows, Blind source separation}

\vskip 0.3in
]

\printAffiliationsAndNotice

\begin{abstract}
In this paper, we show that \textit{slow feature analysis} (SFA), a common time series decomposition method, naturally fits into the \textit{flow-based models} (FBM) framework, a type of invertible neural latent variable models. Building upon recent advances on blind source separation, we show that such a fit makes the time series decomposition identifiable.
\end{abstract}

\section{Introduction}
Data \textit{blind source separation} (BSS) consists in identifying and extracting the factors from which observed data was generated. The objective is to find meaningful information from data in order to help downstream human or machine learning tasks. 

More precisely, identifying components of a (discrete) time series $X \in \mathbb{R}^{T \times d}$ consists in finding an \textit{invertible} function $f$ and some latent signals $S \in \mathbb{R}^{T \times d}$ such that $X=f(S)$. We call $f$ the \textit{mixing} function and each latent variable $S^{(i)}$ $\forall i \in \llbracket 1, d \rrbracket$ is called a \textit{source factor}. The couple $(f,S)$ is a \textit{latent variable model} (LVM) for $X$. The objective of BSS is to identify the true factors $S$ or the function $f$. 

Yet, the full problem is ill-posed, since there exists an infinite number of possible decomposition \cite{hyvarinen1999nonlinear}. Additional assumptions are required to identify the latent factors. First, the latent signals must be independent. The decomposition into independent factors is known as \textit{independent component analysis} (ICA). When $f$ is linear, linear ICA solves the BSS problem under certain assumptions on the sources. Standard assumption for general data is that that sources are non-Gaussian \cite{hyvarinen2000independent}. In the nonlinear general case, assumptions on the source distribution must be coupled with assumptions on the mixing function $f$. Without prior information on the mixing function $f$, we must use an universal approximation function (e.g. a neural network) coupled with a strong assumption on the source distribution. 

Recently, data decomposition using neural LVMs trained by likelihood maximization showed encouraging results \cite{tschannen2018recent}. In particular, the recent work \cite{khemakhem2019variational} gives sufficient conditions of identifiability of the hidden factors estimated with a neural generative model. A version of this work proposed in \cite{sorrenson2020disentanglement} uses \textit{flow-based models} (FBMs) \cite{papamakarios2019normalizing}, a particular type of \textit{invertible} neural networks, to decompose the data. 

In this paper, we build upon \cite{sorrenson2020disentanglement} and focus on a specific assumption for time series decomposition: the \textit{temporal consistency} of the sources, also called \textit{slowness}. Our contribution consists in showing how slowness simply fits in FBM framework (we name it \textit{slow FBMs}), bringing in FBMs the conditions of identifiability for time series source separation. We propose experiments to show that using slow FBMs offers sufficient conditions to identify time series sources. 

We first introduce the notion of slowness in time series decomposition. Then, we present the FBM framework, that can learn data decomposition by exact maximum-likelihood estimation, and show how slowness simply fits in. Finally, after a review of the relation between slowness and BSS, we propose experiments to illustrate the interest of using slow-FBMs instead of FBMs for time series source decomposition. 

\section{Slowness in times series decomposition} 

\textit{Slowness} is a common temporal structure used in time series decomposition. It represents the fact that two consecutive time-steps in a time series have close values. The slower the time series, the closer the consecutive values. The decomposition of time series into consistent (i.e. slow) features is called \textit{Slow feature analysis} (SFA) \cite{wiskott2002slow}. 

We first introduce some notations. We note $Z$ the estimated factors from time series data $X$ such that $S$ is said to be \textit{identified} if there exists $A \in \mathbb{R}^{d \times d}$ such that $S=AZ$. For a time series variable $Z \in \mathbb{R}^{T \times d}$ we note $\langle Z \rangle = \frac{1}{T} \int_{0}^{T}{Z_tdt}$ the mean of $Z$ with respect to time and $\Delta$ the operator of \textit{temporal differentiation} $\Delta Z_{t} = Z_{t} - Z_{t-1}$. We note $\Delta Z = \left[\Delta Z^{(1)} \dots \Delta Z^{(d)} \right]^T \in \mathbb{R}^{T \times d}$ the set of increment signals. Without loss of generality, we impose $Z_0=0$ such that $\Delta Z_1 = Z_1$ is defined.

The slowness is generally seen as the fact that temporal increments of a time series have low variance. The SFA problem is then defined as follows:

\begin{align}
    \min_{f:X=f(Z)} \sum_{i=1}^d \langle \Delta {Z^{(i)}}^2 \rangle \text{ s.t. } \langle Z \rangle = 0, ZZ^T=I
\label{eq:sfa_objective}
\end{align}

Constraints avoid trivial constant solutions and information redundancy between the latent factors. It is also standard but not required to add an ordering constraint for the slowness, i.e. $\langle \Delta {Z^{(i)}}^2 \rangle < \langle \Delta {Z^{(j)}}^2 \rangle$ if $i<j$. 

\paragraph{Standard SFA solving} Standard approach consists in building a set of predefined functions $\tilde{h} = [\tilde{h}_1, \dots, \tilde{h}_K]^T$ where $\tilde{h}_k: \mathbb{R}^d \rightarrow \mathbb{R}$. Then the multivariate signal $\tilde{h}(X)$ is whitened with matrix $W$ and we note $h$ the whitened signal. In standard approaches a PCA is performed on signal $\Delta h(X)$ to extract the slow features \cite{wiskott2002slow}. The main problem comes from the definition of $\tilde{h}$. In simple approaches, $\tilde{h}$ is a set of temporal lags, monomials and pairwise interactions between variables. When the problem is highly non-linear, when there are complicated interactions between variables or when the problem is high-dimensional, this approach is too weak. Enhancing the expressiveness of $\tilde{h}$ by expanding the list of possible functions is not tractable. A way to be exhaustive in $\tilde{h}$ with a small $K$ is to directly optimize in a general function space $\mathcal{H}$. Two highly flexible function spaces are commonly used in machine learning: RKHS \cite{bohmer2011regularized} and neural networks \cite{schuler2018gradient,pfau2018spectral,du2019unsupervised}. 

\paragraph{Maximum-likelihood SFA} We can simply transform SFA into LVM by introducing a probabilistic model for the estimated sources. Constraints in (\ref{eq:sfa_objective}) can be replaced by $Z_t \sim \mathcal{N}(0, I)$ $\forall t \in \llbracket1, T \rrbracket$. Objective (\ref{eq:sfa_objective}) can be replaced by:

\begin{align}
    Z_t | Z_{t-1} \sim \mathcal{N}(Z_{t-1}, I)
    \label{eq:prior_lag}
\end{align}

or equivalently $\Delta Z_t \sim \mathcal{N}(0, I)$ $\forall t \in \llbracket1, T \rrbracket$. In this probabilistic framework, we can solve the SFA problem by maximum-likelihood \cite{turner2007maximum}. In particular, we focus on a recently proposed model to learn data invertible decomposition by maximum-likelihood estimation: the \textit{flow-based models} (FBMs) \cite{papamakarios2019normalizing}. 

\section{Slowness in flow-based models}

\paragraph{Flow-based models}

FBMs are generative models that transform random variable $Z$ with simple distribution into random variable $X$ with complex distribution, with a one-to-one transform. In particular, FBMs are neural LVM, i.e. a neural network $f_\phi$ with parameters $\phi$ that relates latent variables $Z$ to data $X$. We note $f_\phi$ this neural network with parameters $\phi$. In particular, FBMs are \textit{exact likelihood} neural LVM. It means that the parameters $\phi$ can optimized by directly maximizing the likelihood of the data. To achieve this property, while the likelihood of the data is unknown (and generally untractable), FBMs lean on \textit{normalizing flows} (NFs) \cite{rezende2015variational}. A NF is a chain of \textit{invertible} neural networks $\left\{f_{\phi_r} \right\}_{r=1}^R$ with respective parameters $\phi_r$, that enables to pass from a simple distribution $p(Z)$ (e.g. Gaussian) to a complex distribution $p(X)$ (e.g. data) via the change of variable $X = f_{\phi_1} \circ \dots \circ f_{\phi_R}(Z)=f_\phi(Z)$, with $\phi=\left\{\phi_r \right\}_{r=1}^R$. Since transformations are invertible, using the change of variable formula we can define the density of the data $p(X)$:

\begin{align}
    p\left(f_\phi^{-1}(X) \right) \prod_{r=1}^R \left| \det \frac{\partial f_{\phi_r}(x)}{\partial x}\bigg\rvert_{x=f_{\phi_r} \circ f_\phi^{-1} \left(X \right)} \right|^{-1}
\label{eq:normalizing_flows}
\end{align}

\noindent where $f_\phi^{-1}(X)=Z$ and $p(Z)$ is chosen to be simple and explicit. The transformations $f_{\phi_r}$ can be arbitrarily complex as soon as they are invertible. From (\ref{eq:normalizing_flows}), we see that it is required that NFs are easily invertible and have Jacobian determinant easy to compute. 

The idea of using FBM to decompose data into independent components has emerged in \cite{dinh2014nice}. They introduced neural normalizing flows to create a one-to-one generative model that decomposes data into independent Gaussian variables. Contrary to variational autoencoders (VAE) \cite{kingma2013auto}, that also decomposes data into independent Gaussian variables, the FBMs enable to directly maximize the likelihood of the data to learn the neural networks parameters. Finally, the invertibility lets off the necessity to train a decoder that maximizes the likelihood of the latent space representation with respect to data, since FBMs preserve the full information about data in the latent space since they are one-to-one.

\paragraph{Slow flow}

We have seen that NFs are simple invertible transformations whose Jacobian determinant is easy to compute. In particular, the previously introduced temporal differentiation operator $\Delta$ matches these properties. We note $\tilde{Z}=\Delta Z$. We now learn a decomposition $Z$ of $X$ by maximizing the likelihood of $\tilde{Z}=\Delta \circ f_\phi^{-1}(X)$ under the assumption (\ref{eq:prior_lag}) stating that $\tilde{Z}_t \sim \mathcal{N}(0, I)$. We note that the inverse of $\Delta$ is iteratively defined: $\Delta^{-1} \tilde{Z}_t = \tilde{Z}_t + \Delta^{-1} \tilde{Z}_{t-1}$. Hence, $\Delta^{-1}$ can be added as the first NFs of the chain $f_\phi$:

\begin{align*}
    X = f_\phi(Z) = f_\phi(\Delta^{-1} \tilde{Z})
\end{align*}

\noindent Since the Jacobian determinant of the differential operator is simply $1$ (it is a \textit{volume-preserving} NF), adding slowness in time series decomposition with FBM requires only to multiply by the prior density $p(\tilde{Z})=p(\Delta \circ f_\phi^{-1}(X))$ in (\ref{eq:normalizing_flows}). 

We remark that in SFA, the decorrelation between inferred latent factors is obtained by whitening the latent representation. Yet, for example when time series is decomposed into multiple samples, the slowness may change from one sample to the other. Hence, on the one hand, whitening each sample is not appropriate. On the other hand, we found that computing and applying a global whitening matrix for all samples causes unstable training. Using FBMs is a natural solution to keep all the information from data to latent space without explicitly forcing the variance of each latent dimension to be strictly positive. 

Now that we have a flow-based SFA model, the question is: \textit{can we identify the sources with slow-flows?} To answer this question, we need a short review of the literature treating the relation between slowness and identifiability in time series decomposition. 

\section{Relation between slowness and identifiability}

The literature about SFA and BSS is wide and has regularly crossed to build new theories, methods and results on time series decomposition. In this section, we briefly review these crossroads.

We have mentionned above that BSS requires assumptions about the structure of hidden sources. First, without prior knowledge, the sources are assumed independent, such that ICA is the main BSS approach. In linear ICA, a common assumption is the non-Gaussianity of the sources \cite{hyvarinen2000independent}. When data is time series, an alternative assumption is the slowness \cite{hyvarinen2001methods}. In fact, if $Z$ are slow uncorrelated features (see SFA problem above) then $\forall t,i$, $Z^{(i)}_t \approx Z^{(i)}_{t-1}$. Hence, the covariance between $\langle Z^{(i)}_{1:T-1} Z^{(j)}_{2:T} \rangle \approx \langle Z^{(i)}Z^{(j)} \rangle = \delta_{ij}$. It is shown in \cite{tong1991indeterminacy} that having diagonal instantaneous and lagged covariance is a sufficient condition for independence. 
 
In \cite{blaschke2007independent}, they propose a nonlinear BSS method by coupling linear ICA with nonlinear SFA. In \cite{sprekeler2014extension}, they enrich the SFA-based nonlinear ICA with an iterative extraction of the sources. The later showed that if the sources have different autocorrelation functions, then the true sources are identifiable. It constitutes the first theoretically grounded nonlinear BSS. In \cite{hyvarinen2017nonlinear}, they propose a stronger identifiability condition than \cite{sprekeler2014extension} by transforming the sought temporal consistency into a contrastive learning problem \cite{baldi1991contrastive}. They train a neural network to embed time series $X$ such that a classifier can discriminate $[X_{t-1},X_t]$ from $[X_{t^*},X_t]$, with $t^*$ a random time index. They show that, asymptotically and using the universal approximation power of neural networks, the true sources are in the final hidden layer of the neural networks. This classification-based ICA was previously proposed for nonstationary time series in \cite{hyvarinen2016unsupervised} and extended to general ICA in \cite{hyvarinen2019nonlinear}. 

Subsequently, \cite{khemakhem2019variational} proposed a similar identifiability proof, but using maximum-likelihood estimation instead of a contrastive estimation. In particular, they show that using a conditionally factorial prior $p(Z_t | U)$ for the sources instead of a simple factorial prior $p(Z_t)$ is a condition for identifiability. To explicitly extract the estimated sources, they pick $p(Z_t | U)$ into the exponential family of distributions, where the components are respectively functions of $Z$ and $U$. Finally, in \cite{sorrenson2020disentanglement}, they use FBMs to complete the last-mentioned identifiability proof with truly invertible functions. In theory, this auxiliary variable $U$ can be for example $Z_{t-1}$, finding back the notion of slowness. Yet, this assumption is not experimented in the aforementioned papers. 

\paragraph{Identifiability in slow-FBMs} We therefore can use the recent results of \cite{khemakhem2019variational,sorrenson2020disentanglement} to understand that the estimated features with slow-FBMs are the true sources up-to linear demixing. In particular, in the limit of infinite data and with respect to the Gaussian assumptions of SFA, after training we have $S=A f_\phi^{-1}(X)$. Matrix $A$ is generally found by solving the linear ICA of the invertible embedding $Z=f_\phi^{-1}(X)$. Hence, adding a simple slow flow at the beginning of the chain of normalizing flows is a sufficient condition for the separation of source signals.

\section{Experiments}

In this section, we propose two simple experiments to show how adding a slow-flow in a FBM enables time series source separation.  

\subsection{Setup and implementation details}

In both experiments, we compare slow-FBM with standard FBM (the only difference is the absence of the slow flow) and with linear ICA (to verify that the problem is not linearly identifiable). We do not compare to methods cited in the related work since we want to focus on the main point of this short paper, that is: the slowness is a way to induce identifiability in time series decomposition. 

We use the RealNVP \cite{dinh2016density}, a standard flow-based model, as the invertible embedding function $f_\phi$.

\subsection{Decomposition into structural components}

\begin{figure}[h]
  \centering
  \includegraphics[width=0.32\linewidth]{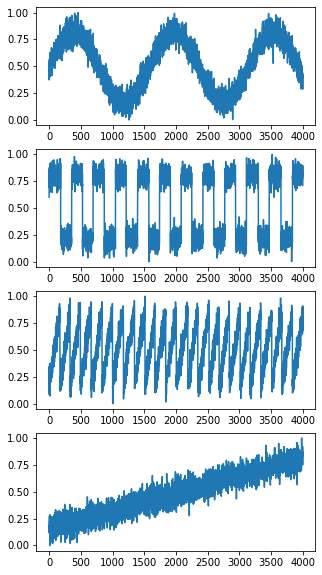}
  \includegraphics[width=0.32\linewidth]{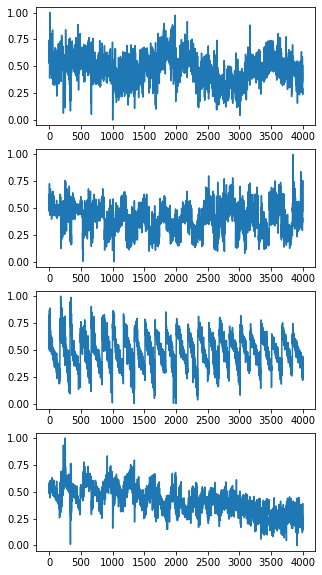}
  \includegraphics[width=0.32\linewidth]{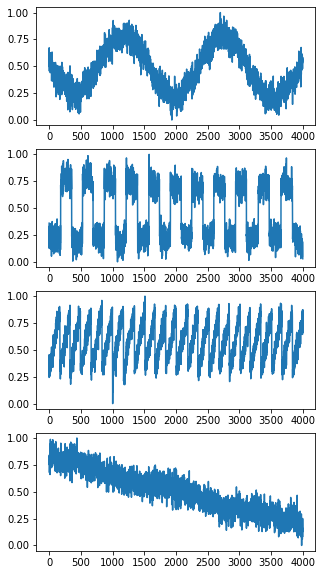}
  \caption{BSS of mixed structural components. \textbf{Left}: Ground truth independent components. \textbf{Middle}: Estimated components with standard FBM. \textbf{Right}: Estimated components with Slow-FBM. The maximum absolute mean correlations with true signal are given in Table \ref{tab:structural_results}.}
  \label{fig:ICA_structural}
\end{figure}

In this first experiment, we use a common example: the decomposition of time series into \textit{structural components}. The structural components consist in trends, cycles, the seasonality and the irregularity. The trend is a monotone long term signal. The cycle is a signal that affects the observations at random frequencies. Seasonality is a signal that affects the time series at fixed frequency. The exogenous input can be a control variable, an environment factor or some random noise. We generate one seasonal signal, two asynchronous cycles and one trend. These four components are corrupted by additive Gaussian noise $\mathcal{N}(0, 0.2)$ (see left plot in Figure \ref{fig:ICA_structural}). The four components are assumed independent. 

For the experiment, we mixed the sources $S$ with a randomly initialized FBM.

\begin{table}[h]
    \centering
    \begin{tabular}{|C{0.08\textwidth}C{0.1\textwidth}C{0.08\textwidth}C{0.12\textwidth}|}
    \hline
     ICA & FBM + ICA & S-FBM & S-FBM + ICA  \\
    \hline
     $ 58.2 \pm 7.7 $   & $ 66.2 \pm 3.7 $  & $ 79.6 \pm 3.1 $ &   $ \textbf{94.3} \pm 3.1  $     \\
    \hline
    \end{tabular}
    \caption{Maximum absolute correlation between estimated and true sources. S-FBM stands for slow-FBM.}
    \label{tab:structural_results}
\end{table}

The results clearly show that adding slowness into the FBM model enables to identify the sources. Moreover, we can see that even before the final demixing linear ICA, the slow-FBM gives reasonably good decomposition of the data. 

\subsection{Audio demixing}

We propose a second standard experiment. We propose to recover mixed audio samples. We chose four random instrumental samples and mixed them with a randomly initialized FBM. First $75 \%$ of the signal serves as train set, the rest as test set. The four components are assumed independent. 

\begin{figure}[h]
  \centering
  \includegraphics[width=0.32\linewidth]{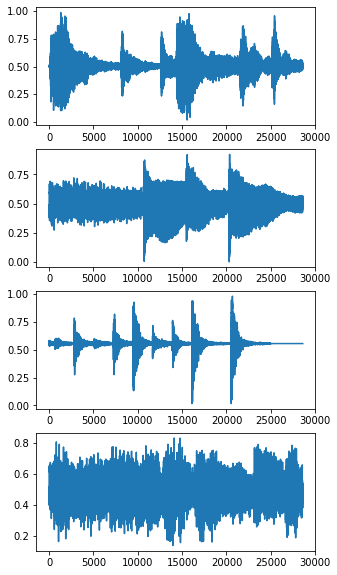}
  \includegraphics[width=0.32\linewidth]{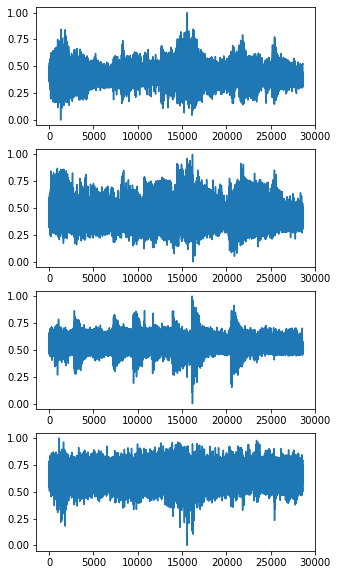}
  \includegraphics[width=0.32\linewidth]{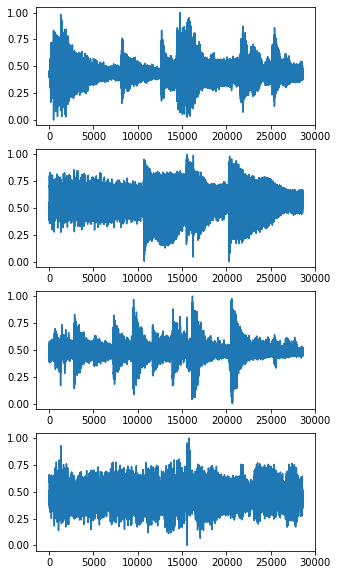}
  \caption{BSS of mixed audio, on test sample. \textbf{Left}: Ground truth independent components. \textbf{Middle}: Estimated components with standard FBM. \textbf{Right}: Estimated components with Slow-FBM. The maximum absolute mean correlations with true signal are given in Table \ref{tab:audio_results}}
  \label{fig:ICA}
\end{figure}

We observe that the decomposition with slow-FBM is noisy, in particular for the sparse signal (number 3 from top). This is due to the little size of the data we used. Adding more data improves the identifiability of true sources (asymptotic result in \cite{khemakhem2019variational}). 

\begin{table}[h]
    \centering
    \begin{tabular}{|C{0.08\textwidth}C{0.1\textwidth}C{0.08\textwidth}C{0.12\textwidth}|}
    \hline
     ICA & FBM + ICA & S-FBM & S-FBM + ICA  \\
    \hline
    $ 52.0 \pm 1.7 $     & $ 68.8 \pm 1.7 $ & $ 77.4 \pm 0.9 $ & $ \textbf{95.7} \pm 3.4 $ \\
    \hline
    \end{tabular}
    \caption{Maximum absolute correlation between estimated and true sources. S-FBM stands for slow-FBM.}
    \label{tab:audio_results}
\end{table}

The results again clearly show that adding slowness into the FBM model enables to identify the sources. Moreover, we can see that even before the final demixing linear ICA, the slow-FBM gives reasonably good decomposition of the data. 

\section{Conclusion and perspectives}

In this short paper, we analyzed a simple way to induce identifiability in time series decomposition with flow-based models using slowness. In particular, we used the fact that temporal differentiation is an invertible transform that can be plugged in a chain of normalizing flows. We then related the addition of slowness into FBM to the recent nonlinear blind source separation methods to finally show in simple experiments the advantage of using the slow-FBM instead of FBM for time series decomposition. 

\paragraph{Acknowledgments} This work is supported by the company Safran
through the CIFRE convention 2017/1317.

\bibliography{main_inn}
\bibliographystyle{icml2020}

\end{document}